\documentclass{article}

\usepackage{microtype}
\usepackage{graphicx}
\usepackage{subfigure}
\usepackage{booktabs} 
\usepackage{amsmath} 
\usepackage{tikz}
\usetikzlibrary{bayesnet}

\usepackage{hyperref}

\usepackage[ruled,linesnumbered]{algorithm2e}
\usepackage{tabularx}
\usepackage{caption}
\usepackage{enumitem}
\usepackage{titlesec}

\usepackage[accepted]{icml2021}

\icmltitlerunning{Kernel Learning for Mobile Health}
\begin{document}
\twocolumn[
\icmltitle{Online structural kernel selection for mobile health}


\begin{icmlauthorlist}
\icmlauthor{Eura Shin}{harvard}
\icmlauthor{Pedja Klasnja}{umich}
\icmlauthor{Susan Murphy}{harvard}
\icmlauthor{Finale Doshi-Velez}{harvard}
\end{icmlauthorlist}

\icmlaffiliation{harvard}{Department of Computer Science, Harvard University}
\icmlaffiliation{umich}{School of Information, University of Michigan}

\icmlcorrespondingauthor{Eura Shin}{eurashin@g.harvard.edu}

\icmlkeywords{Compositional kernel search, mobile health, online learning, multitask learning}

\vskip 0.3in
]
\icmlkeywords{Mobile health, compositional kernel search, multi-task learning}

\printAffiliationsAndNotice{} 

\begin{abstract}
Motivated by the need for efficient and personalized learning in mobile health, we investigate the problem of online kernel selection for Gaussian Process regression in the multi-task setting. We propose a novel generative process on the kernel composition for this purpose. Our method demonstrates that trajectories of kernel evolutions can be transferred between users to improve learning and that the kernels themselves are meaningful for an mHealth prediction goal. 
\end{abstract}

\section{Introduction}
Smart phones permeate every aspect of our modern lives; this makes them the perfect platform for delivering tailored healthcare solutions.  Today, mHealth solutions assist patients with managing weight, mental health, medication without the confines of a clinic or a structured outpatient program.  In this work, we will focus on HeartSteps, a physical activity app, that assists individuals with Stage 1 hypertension with adoption of regular physical activity \cite{klasnja2019efficacy}.  To send contextually-tailored activity reminders, it is useful to learn the relationship between different contexts (e.g. sunny weather, morning time) and the user's activity. Our goal is to predict the user's activity as a function of context and the intervention. 

MHealth is a challenging learning environment because user-level data is low volume, noisy, and high dimensional. Gaussian Process (GP) regression is a suitable model for this, as its uncertainty estimates are useful in guiding downstream decision making \cite{tomkins2019intelligent} and it provides an interpretable generalization of familiar linear regression models.  The choice of kernel is critical to the performance of a GP regression model, and the best kernel depends on the application and data at hand.
\begin{figure}[h]
    \centering
    \includegraphics[width=.8\linewidth]{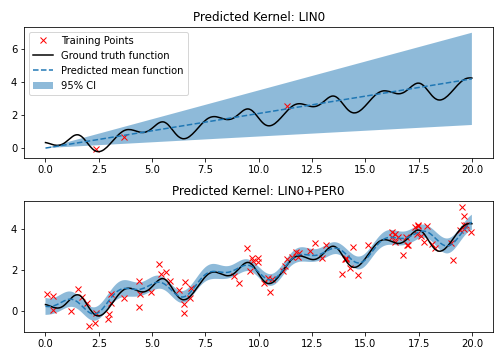}
    \caption{The kernel recovered for 1-D regression task with $3$ training points (top) and $187$ training points (bottom). $\mathrm{LIN0}$ and $\mathrm{PER0}$ refers to a linear and periodic kernel on the first (and only) dimension of the data, respectively. This synthetic experiment demonstrates our model's ability to infer the kernel composition for an unseen task and to adjust the composition with growing data.}
    \label{fig:toy functions visualized}
\end{figure}

Some work has attempted to automate the kernel selection process by selecting a compositional kernel from the data \cite{steinruecken2019automaticStatistician, duvenaud2013cks}. Compositional kernels are defined by the sum and product of simpler kernels. Oftentimes, each kernel component corresponds to a meaningful representation of a covariate in the data; for example, a periodic kernel on the time may indicate that a user's physical activity fluctuates periodically with respect to the time of day. This additive structure is more interpretable \cite{schulz2016probing, lloyd2014ABC} and has better generalization in high dimensional data \cite{duvenaud2011additive, duvenaud2013cks} compared to the popular squared-exponential automatic relevance determination (SE-ARD) kernel.

A promising yet unexplored application of compositional GPs is the online data setting. Here, we can add kernel components as a way to increase the complexity of our model, which means that the model's representation power can grow as the data and its observable structures grows. This model's behavior is desirable in the mobile health regime, where data is continuously collected through interactions with the user and the model should adapt accordingly.

Unfortunately, kernel selection with batch data is already a difficult open research problem, and 
current methods do not offer an incremental approach for kernel selection. To apply current methods to online data, for each new example, one must update the batch data set and re-run the entire selection algorithm. 
Current approaches are also sensitive to the local optima induced by the kernel hyperparameters and require multiple random restarts \cite{duvenaud2013cks, kim2018scaling} or duplicitous kernels \cite{tong2020shrinkage, sun2018differentiable} for correct model comparison. 

We address these challenges by transferring information about the kernel composition between users (tasks). 
In this paper we develop a generative process to explicitly capture how kernel compositions evolve with online data. Furthermore, we explore the idea that these trajectories of kernel evolution can be \emph{transferred} between users. 
We demonstrate that our model can leverage past data from other kernel trajectories to perform kernel selection on new data without requiring any retraining, that learning group-level kernel trajectories has comparable performance to the completely user-level baselines while improving speed and interpretability, and that our approach is effective on a real world application in mobile health.

\section{Related Work} 
\paragraph{Compositional Kernel Selection}
There is a rich body of literature on the Automatic Statistician, where the goal is to extract an appropriate kernel function from data by searching over a grammar of compositions \citep{steinruecken2019automaticStatistician}. Work by \citet{duvenaud2013cks} and \citet{lloyd2014ABC} introduced Compositional Kernel Search (CKS), a greedy search over the sums and products of simpler base kernels to maximize the model evidence. Since the search space for this method is combinatorially large, later work has focused on making CKS scalable by approximating the model evidence \cite{kim2018scaling} or representing the search as a neural network \cite{sun2018differentiable}. 

Instead of considering kernel learning as a discrete search problem, another line of work has taken a more probablistic approach. \citet{malkomes2016bayesian} developed a "kernel kernel" to perform Bayesian optimization for kernel selection. \citet{gardner2017hmcBayesOpt} searches over the space of \emph{additive kernels} by finding a subset of squared-exponential kernels on independent features, but does not consider different structural forms (linear, periodic). Finally, \citet{titsias2011spike, tong2020shrinkage} take an approach inspired by feature selection and use shrinkage priors to select a subset of kernels for the composition. 

None of these approaches explicitly build the streaming nature of the data into their model. The greedy search method by \citet{duvenaud2013cks} is similar to our own, in that it prioritizes building the strongest signals (kernels) from the data into the composition. However, it is computationally infeasible for our applications, as it would require restarting the entire search process from scratch each time new data arrives. More suitable to the application is the Bayesian model from \citet{tong2020shrinkage}, which allows the previous posterior to serve as the new prior, but this completely task-level selection algorithm still requires reoptimizing for the posterior with every batch update. In contrast, we propose a generative process for kernel compositions over time, which transfers information between users about how a kernel will evolve in complexity as data accrues.

\paragraph{Multi-task learning (MTL) with Gaussian Processes}
The MTL scenario involves learning a separate Gaussian Process function for each of the related tasks. Previous work has considered sharing information between these tasks on many different levels of the inference. On the parameter level, \citet{yu2005learningGPMultiple} uses the weight-space view of Gaussian Processes to derive a kernel function which relates multiple tasks, while \citet{tomkins2019intelligent} and \citet{luo2018mixedEffectsInjection} extend the mixed-effects models from Bayesian linear regresison to GP kernels. Rather than transferring ``model parameters," \citet{srijith2014jointMultitaskARD} and \citet{leroy2020magma} transfer the ``kernel hyperparameters", such as the lengthscales or amplitude parameters. Finally, on the model level, \citet{tong2019latentCovarianceMultipleTimeseries} and \citet{titsias2011spike} transfer the commonalities of the structure between functions by sharing the kernels themselves, while \citet{li2018hierarchicalMultitask} learn a mixture of Gaussian Processes on a lower dimensional latent space. Instead of transferring knowledge about the model, we transfer knowledge about the model's evolution \emph{over time} to conduct \emph{trajectory level} transfer between tasks. 

\paragraph{Online Models in Mobile Health}
Due to the challenges of modeling human behavior in a limited data setting, it is natural to use (contextual) bandit algorithms with simple models for the reward function.
For example, in HeartSteps, an app designed to increase  physical activity, the authors use Bayesian linear regression to model the proximal outcome (the number of steps following a decision point) \citep{liao2020heartsteps}. 
However, the model makes the unrealistic assumption that the reward is linear with respect to the state features. Similar to our work, \cite{tomkins2019intelligent} reformulate their reward function as a Gaussian Process but still elect to use a simple linear kernel.

\section{Background}
\paragraph{Compositional Kernels}
We assume that the reader is familiar with Gaussian Process regression \citep{books/lib/RasmussenW06}. Compositional kernels express complex functions through the sum and product of simpler base kernels \cite{duvenaud2013cks}. A base kernel is a kernel function, such as the linear ($\mathrm{LIN}$), squared-exponential ($\mathrm{SE}$), and periodic ($\mathrm{PER}$), applied to a single covariate in the data. These kernels are related through the fact that for independent $f_1 \sim GP(0, k_1)$ and $f_2 \sim GP(0, k_2)$, the sum and product of the functions corresponds to the same operations on the kernels. That is,  $f_1 + f_2 \sim GP(0, k_1 + k_2)$ and $f_1 \times f_2 \sim GP(0, k_1 \times k_2)$. 

Compositional kernel selection involves combining these rules to form an expressive and readily interpretable grammar over functions. For example, some popular functions that can be found via compositional kernel selection include Bayesian polynomial regression ($\mathrm{LIN} \times \mathrm{LIN} \times \ldots \times \mathrm{LIN}$), generalized additive models ($\sum_{d=1}^D \mathrm{SE}_d$), and automatic relevance determination ($\prod_{d=1}^D \mathrm{SE}_d$).

\paragraph{Dirichlet Process Mixture Models}
A Dirichlet Process (DP) mixture model places a prior over mixtures with an infinite number of components.  The generative process is:
\begin{align}
    &G \sim DP(\alpha,H_0)\\
    &Z_n | \pi \sim \mathrm{Mult}(\pi) \\
    &X_n | Z_n \sim p(X_n ; \theta_{Z_n} )
\end{align}
where $X_n$ is the observation, $Z_n$ is the cluster assignment and $G = \{ \pi , \{ \theta_j \} \}$ is the mixture model that is the draw from the Dirichlet process; the weights $\pi_j$ are the probability of each mixture component and $\theta_j$ are the parameters of each component.  The concentration parameter $\alpha$ governs the number of expected components in a finite sample, and $H_0$ is the base distribution over mixture parameters.  For more information on Dirichlet processes, see \citep{teh2010dirichlet}.

It is possible to track the cluster assignments 
under this DP using a Chinese Restaurant Process (CRP). 
Given the current cluster assignments $\mathbf{Z} = Z_1, \ldots, Z_{n - 1}$ to $K$ unique clusters, the $n$-th customer is assigned to a partition $k$ with the following probability: 
\begin{align}
    \label{eq: CRP probabilities}
    p(Z_{n} = k | \mathbf{Z}) = 
    \begin{cases}
    \frac{\alpha}{N + \alpha - 1},& \text{if } k = K + 1\\
    \frac{N_k}{N + \alpha - 1},              & \text{otherwise}
    \end{cases}
\end{align}
where $N_k$ is the number of points already assigned to partition $k$. Under this model, the probability of starting a new partition is determined by $\alpha$. The probability of being assigned to an existing partition is proportional to the number of points already within that partition. 

\begin{figure}[t]
    \centering
    \includegraphics[width=1\linewidth]{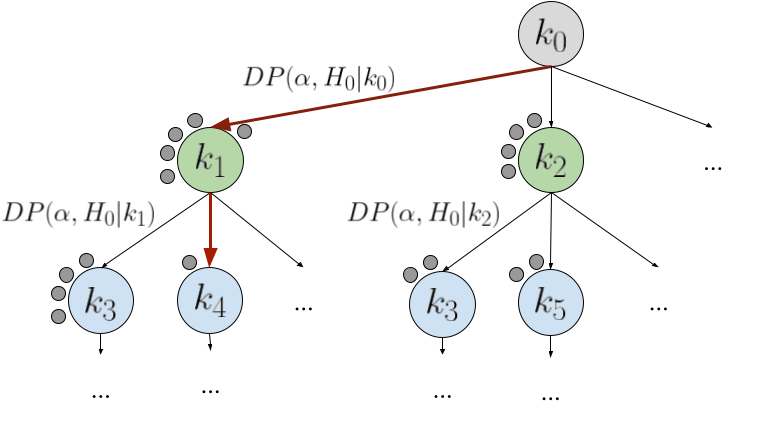}
    \caption{An example trajectory model. Nodes represent kernel compositions and edges the transitions from one composition to the next after a single time step. Customers (data at time step $t$ for user $m$) are associated with each node, representing the number of times a kernel transitioned from the relevant parent to child node. 
    All non-leaf kernels have a Dirichlet Process prior on the outgoing edges, allowing the tree to \lq\lq branch" with the data. The prior on all non-leaf nodes is the same. An example trajectory through the model structure is bolded in red. \vspace{0.3em}}
    \label{fig:kernel evolution tree}
\end{figure}

\section{Methods}

\subsection{Problem Setup} 
There are $M$ total users.  Data for each user arrives online in batches.  We denote $N_{m, t}$ as the total number of observations so far for user $m$ at time $t$, and the total number of batches for user $m$ as $T_m$.  Each batch may have a different number of observations. In the following, we will be considering what kernel composition best models \emph{all} of the data we have up to time $t$ from user $m$.  For this, let $\mathcal{D}_{m, t} = \{\mathbf{X}_{m, t}, \mathbf{y}_{m, t}\}$  represent the \emph{cumulative} data set of all observations for user $m$ up to time $t$. Here, $\mathbf{X}$ are the contexts and $\mathbf{y}$ are the targets (e.g. step-count).
That is, for user $m$ we are interested in finding a trajectory of $T_m$ kernels: 
\begin{align}
\label{eq: kernel trajectory}
    \mathcal{T}_m = \left \{K_{m, 1}, K_{m, 2}, \ldots, K_{m, T_m}\right \}
\end{align}
Each of the kernels in Equation \ref{eq: kernel trajectory} have corresponding hyperparameters $\boldsymbol{\theta}_{m, 1}, \ldots, \boldsymbol{\theta}_{m, T_m}$.

Importantly, we are assuming a stationary setting where the \emph{ground truth kernel for each user is fixed} and does not change over time. Given this assumption, the evolving kernels in the trajectory do not represent an evolving ground truth, but rather, our model's \emph{adjustments to the bias-variance tradeoff}. 

Moreover, our model exploits the common kernel transitions that exist among different users. For example, if we have observed many users undergo the transition from $K_t = k_\mathrm{linear}$ to $K_{t + 1} = k_\mathrm{linear} + k_\mathrm{periodic}$, then a user with a $k_\mathrm{linear}$ kernel is more likely to transition into the $k_\mathrm{linear} + k_\mathrm{periodic}$ kernel at the next time step, as opposed to a $k_\mathrm{se} + k_\mathrm{periodic}$ kernel, a notion we formalize in the next section. 

\subsection{Generative Process}
We represent the evolution of kernel compositions from $K_{t}$ to $K_{t + 1}$ as a Chinese Restaurant Process. For simplicity, let us refer to these evolutions as transitions of the form $K_\mathrm{current} \rightarrow K_\mathrm{next}$. Critically, the distribution over $K_\mathrm{next}$ is determined by the originating composition $K_\mathrm{current}$, so our model assumes a \textbf{separate restaurant per unique $K_\mathrm{current}$}. This means that our trajectory model is \emph{not hierarchical}, but tracks a separate transition for every unique parent kernel.

Each restaurant represents the Dirichlet Process over realizations of $K_\mathrm{next}$ when coming from $K_\mathrm{current}$. Each table in the restaurant corresponds to a possible composition that $K_\mathrm{next}$ could manifest and encodes a distribution over functions (a GP parametrized by $K_\mathrm{next}$). Customers are data sets $\mathcal{D}_{m, t}$ at time point $t$ for user $m$. When a customer sits at a table, $K_{m, t}$ assumes the composition represented at that table and contributes a \emph{count} toward that composition. 

It is worth noting some of the implications of this Chinese Restaurant Process. We assume that most kernels coming from $K_\mathrm{current}$ will evolve into a previously observed $K_\mathrm{next}$. There is always room for exception; there is a chance that $\mathcal{D}_{m, t}$  will be so different that it is seated at a new table (i.e. $K_{m, t}$ is not represented any of the current tables). The customers $\mathcal{D}_{m, t}$ are exchangeable in that the joint distribution (over $K_\mathrm{next}$) is not affected by the order in which the customers arrive. By modeling the most probable transitions among compositions, we are also building a directed graph which represents how kernels relate to one another over time. An example of such a graph is in Figure \ref{fig:kernel evolution tree}.

As motivated by our analogy, for every kernel composition $K_\mathrm{current}$, there is a separate Dirichlet Process over the next composition $K_\mathrm{next}$: 
\begin{align}
    & H_{K_\mathrm{current}}(K_\mathrm{next}) \sim DP(\alpha, H_{0} | K_\mathrm{current}) \label{eq: model DP} 
\end{align}
where $H_K$ defines a distribution over the next composition leaving from $K$. $H_0$ is the base distribution, and $\alpha$ the concentration parameter of the DP.
The choice of parameters for the experiments can be found in Appendix \ref{appendix: toy priors} and \ref{appendix: hs priors}.  

Recall that we are using this model to govern the bias-variance trade-off in choosing a sufficiently, but not overly, complex kernel composition to fit the data we have about a user so far. The generative process for these cumulative data $\mathcal{D}_{m, t}$ is: 
\begin{align}
    & K_{m, t} \sim H_{K_{m, t - 1}} \label{eq: model dist over kernels}\\
    & \theta_{m, t}  \sim p(\theta_{m, t} | K_{m, t}) \label{eq: model hypers}\\
    & f_{m, t} \sim GP(f_\mu, K_{m, t}) \label{eq: model GP}\\
    & Y_{m, t} \sim \mathcal{N}(f_{m, t}, \sigma_{m, t}^2) \label{eq: model likelihood}
\end{align}

Though in our model $Y_{m, t}$ will overlap in data with $Y_{m, t - 1}$, we use different generative processes for each by noting that the bias-variance trade-off, and therefore the best kernel, may differ between the two partially overlapping data sets. Once the kernel composition is defined, Eq. \ref{eq: model GP} and \ref{eq: model likelihood} mean that the observations are distributed according to a standard Gaussian Process with mean function $f_\mu$, kernel $K_{m, t}$, and signal noise $\sigma_{m, t}^2$. The composition at time step $t$ is sampled from the distribution over kernels defined by the preceding composition (Eq. \ref{eq: model dist over kernels}). We enforce a prior over kernel hyperparameters (Eq. \ref{eq: model hypers}). This enables us to incorporate domain knowledge on quantities such as the periodicity and to sufficiently separate the model classes. For example, a squared exponential kernel with an extremely large lengthscale maps to functions that behave linearly; a prior on the lengthscale would minimize overlap between the Linear and SE kernels.

\subsection{Inference}
Let $\mathbf{K}^*$ represent the set of all allowed kernel compositions represented among all users and time steps. 
The unknown parameters are $\boldsymbol{\Theta} = \left (\boldsymbol{Z}, \boldsymbol{K}, \boldsymbol{\theta}, \boldsymbol{f} \right )$:  
\begin{itemize}[leftmargin=*]
\label{model parameters}
\itemsep 0em
    \item The table assignments for customers at the $K^\mathrm{th}$ restaurant $\boldsymbol{Z}^{(K)} = \left \{Z_{m, t} | K_{m, t - 1} = K \right \}$ for all $K \in \mathbf{K}^*$
    \item The kernel composition at each table, where $C^{(K)}$ represents the total number of occupied tables in the $K^\mathrm{th}$ restaurant $ \boldsymbol{K}^{(K)} = \left \{K_{c} \right \}_{c = 1}^{C^{(K)}}$ for all $K \in \mathbf{K}^*$
    \item The hyperparameters of each kernel $\boldsymbol{\theta}^{(K)}$ for all $K \in \mathbf{K}^*$
    \item The Gaussian Process function for each customer, defined by the kernel composition (with corresponding hyperparameters) at the table $\boldsymbol{f} = \{f_{m, t}\}$ for all $m = 1, \ldots, M$ and $t = 1, \ldots, T_m$
\end{itemize}
DP related parameters are indexed by $(K)$ because there is a separate Dirichlet Process for each kernel composition. 
The kernel composition for a customer depends on the composition plated at the table, such that $K_{m, t} = \mathbf{K}_{Z_{m, t}}^{(K_{m, t-1})}$. Recall that $\mathbf{K}^{(K_{m, t-1})}$ is the set of kernel compositions at each table of the restaurant (DP) conditioned on the kernel $K_{m, t-1}$ and $Z_{m, t}$ is the seating of $\mathcal{D}_{m, t}$. 


\subsection{Optimization}
\label{sec:optimization blocks}
The inference objective is to sample from the joint posterior, $p(\mathbf{Z}^{(K)}, \mathbf{K}^{(K)} | \boldsymbol{\mathcal{D}})$ where $\boldsymbol{\mathcal{D}} = \{\mathcal{D}_{m, t}\} $ is the set of data for all $m$, $t$ available at the time of training the model. 
We implement a stochastic optimization procedure similar to the marginal Gibbs sampler in \cite{gelman2013bayesianDataAnalysis}. Our optimization alternates between the following three steps: 

\begin{enumerate}[leftmargin=*]
\itemsep 0em
    \item \textbf{Seating the customers.} Sampling from the posterior $p(\boldsymbol{Z}^{(K)} | \boldsymbol{K}^{(K)}, \boldsymbol{\mathcal{D}})$; marginalizing out $\boldsymbol{\theta}^{(K)}, \boldsymbol{f}$
    \item \textbf{Plating the tables.} Sampling from the posterior $p(\boldsymbol{K}^{(K)} | \boldsymbol{Z}^{(K)}, \boldsymbol{\mathcal{D}})$; marginalizing out $\boldsymbol{\theta}^{(K)}, \boldsymbol{f}$
    \item \textbf{Approximating the kernel evidence.} Updating the $p(\boldsymbol{\theta} | K)$  according to the current compositions among the data 
\end{enumerate}
 
Steps $(1)$ and $(2)$ are exactly as in the marginal Gibbs sampler. Step $(3)$ allows us to empirically update the distribution over hyperparameters for each kernel. The kernel hyperparameters are critical kernel selection but difficult to specify in advance, especially for a large space of potential kernels. Step (3) allows us to update $p(\boldsymbol{\theta} | K)$ with optimization and is discussed below.

\paragraph{Seating the Customers}
In this section we discuss updates for $\boldsymbol{Z}^{(K)}$. The seating assignment is sampled from a multinomial posterior, with the probability 
\begin{align}
\begin{split}
    &p(Z_{m, t} = c| \boldsymbol{\mathcal{D}}, \mathbf{K}^{(K)}) = \\
    & p(\mathbf{y}_{m, t} | \mathbf{X}_{m, t}, Z_{m, t} = c, \mathbf{K}^-_c)p(Z_{m, t} = c | \mathbf{Z}^-, Z_{m, t-1})
\end{split}
\end{align}
where $\mathbf{K}^-$ and $\mathbf{Z}^-$ are the remaining kernel compositions (at occupied tables) and seating assignments in restaurant $K$ after unseating customer $Z_{m, t}$. The kernel likelihood $p(\mathbf{y}_{m, t} | \mathbf{X}_{m, t}, \mathbf{K}^-_c)$ is calculated as discussed in Section $\ref{sec: kernel evidence}$. The table asssignment probability $p(Z_{m, t} | \mathbf{Z}^-, Z_{m, t-1})$ is as determined by the CRP from Eq. \ref{eq: CRP probabilities}.

\paragraph{Plating the Tables}
In this section we discuss updates for $\mathbf{K}^{(K)}$. We assign a kernel composition to each table by sampling from the posterior distribution, under the prior likelihood for the composition $H_0$ and the likelihood defined by the customers (data) seated at the table:
\begin{align}
\begin{split}
    & p(\mathbf{K}^{(K)}_c | \boldsymbol{\mathcal{D}}, \boldsymbol{Z}^{(K)}) =  \\
    & H_0(\mathbf{K}^{(K)}_c) \times \prod\limits_{m, t : Z^{(K)}_{m, t} = c} p(\mathbf{y}_{m, t} | \mathbf{X}_{m, t},\mathbf{K}^{(K)}_c)
\end{split}
\end{align}

We obtain samples via the Metropolis-Hastings algorithm. The algorithm is intialized at the table's current composition. New compositions are proposed by adding or removing kernels from the current composition, as in Appendix \ref{appendix: mh proposer}.

\paragraph{Approximating the Kernel Evidence}
\label{sec: kernel evidence}
In comparing potential kernel compositions, both the local and global update procedures require calculating the model evidence, $p(\mathbf{y} | \mathbf{X}, K)$, which contains an integral over the latent functions $f$ and hyperparameters $\theta$. Under the GP model, the marginalization of latent functions $f$ is straightforward and yields the equation:

\begin{align}
\label{eq: gp log marginal likelihood}
 \log p(\mathbf{y} | \mathbf{X}, K, \boldsymbol{\theta}) = -\frac{1}{2}  (\mathbf{y}^\top K^{-1} \mathbf{y}  +\log|K| + n\log2\pi)
\end{align}

where $K = K_{\boldsymbol{\theta}}(X, X) + \sigma^2 I$ is the kernel function parametrized by $\boldsymbol \theta$ and evaluated at $n$ input points $X$, with our noise variance $\sigma^2$.  As in most GP schemes, we assume that the noise variance, $\sigma^2$ is a kernel hyperparameter. 

In contrast, the integral over kernel hyperparameters is intractable. We approximate this quantity by taking $S$ samples from $p(\boldsymbol{\theta} | K)$, so that $p(\mathbf{y} | \mathbf{X}, K) \approx \frac{1}{S} \sum \limits_{s = 1}^{S} p(\mathbf{y} | \mathbf{X}, K, \boldsymbol{\theta}_s)$. The quantity $p(\mathbf{y} | \mathbf{X}, K, \boldsymbol{\theta}_s)$ is the GP log marginal likelihood from Eq. $\ref{eq: gp log marginal likelihood}$. In practice, it is difficult to correctly specify the prior over hyperparameters, $p(\boldsymbol{\theta}| K)$ in advance.
As a result, we update $p(\boldsymbol{\theta} | K)$ for every iteration by calculating an empirical distribution over the maximum a posteriori (MAP) optimized hyperparameters for each kernel composition (described in Appendix \ref{appendix: hyperparameter updates}). 

\paragraph{Prediction}
\label{predictions}
Once the model is trained, we are able to select the kernel at a new time step $K_{m, t + 1}$ based on the kernel at the previous time step ($K_{m, t}$). We accomplish this in a greedy manner by selecting the new kernel as one that maximizes the kernel evidence on the updated data set, $\mathcal{D}_{m, t + 1}$. This corresponds to greedily traversing the directed graph (example in Fig. \ref{fig:kernel evolution tree}) with an option to self-loop at every node. Our procedure is given in Algorithm \ref{alg: trajectory prediction} in the Appendix \ref{appendix: hs prediction}. 

\section{Experimental Setup}
\paragraph{Baselines}
Our baselines include the discrete, generative Compositional Kernel Search (CKS) proposed by  \cite{duvenaud2013cks} and the shrinkage model proposed by \cite{tong2020shrinkage}. All experiments have two mutually exclusive subsets of users, the training users $\mathbf{m}_\mathrm{train}$ and testing users $\mathbf{m}_\mathrm{test}$. We report results for all models on $\mathbf{m}_\mathrm{test}$. Since both baselines are intended for batch data, their training procedures differ from our Trajectory model and are outlined in Appendix Table \ref{table: train baselines}. The Trajectory model is only trained on data from users $m \in \mathbf{m}_\mathrm{train}$. 

For testing, all three methods will use $K_{m, t}$ (the kernel at the current time step) to predict on data from the next time step $t + 1$ for user $m \in \mathbf{m}_\mathrm{test}$. The kernel $K_{m, t}$ is always determined by $\mathcal{D}_{m, t}$. Since both of the baselines  use the current users's data for inference, they will \emph{retrain} on $\mathcal{D}_{m, t}$ to test on future data from the same $m \in \mathbf{m}_\mathrm{test}$. The trajectory model will \emph{not require retraining}, but still uses $\mathcal{D}_{m, t}$ to determine $K_{m, t}$ according to the procedure outlined in Algorithm \ref{alg: trajectory prediction}.

\paragraph{Candidate Kernel Pool}
Consistent with previous kernel search methods, we will consider the linear ($\mathrm{LIN}$), periodic ($\mathrm{PER}$), and squared-exponential ($\mathrm{SE}$) kernel functions in forming the compositions. We represent a kernel component by pairing the kernel function with a dimension. For example, $\mathrm{LIN}0$ denotes a linear kernel function which operates only on the $0$-th dimension of the data. 

For CKS, the initial kernel pool is the set of kernel components defined by pairing \emph{every kernel function} to \emph{every dimension}. For $D$ dimensional data, this means there are $D \times 3$ initial kernel components, since we consider $3$ kernel functions. The Shrinkage and Trajectory methods are \emph{additive}, meaning they form compositions by selecting a subset of product kernels to add together \cite{duvenaud2011additive}. Though these methods cover a smaller space of compositions than CKS, the pool of product kernels to choose from is typically large, usually representing all possible products of kernel components up to a fixed order. The candidate kernel pools for the synthetic and Heartsteps experiments are in Appendix \ref{appendix: toy pool} and \ref{appendix: hs pool} respectively. 

\paragraph{Selecting a Trajectory Model}
Each iteration of the stochastic optimization produces a structure for the Trajectory model. Figure \ref{fig:toy trajectory model} is one such structure found during optimization. In Appendix \ref{appendix: toy model alternative}, we provide alternate structures from different points in the optimization. Without knowledge of the ground truth composition and in multi-dimensional settings, different trajectory models may fit the data equally well, as is the case with the Heartsteps experiment. In this scenario one can ask a domain expert to select the structure that is most aligned with domain knowledge. In our Heartsteps experiments, we chose the model with the simplest structure among $5$ different models with similar log likelihoods, shown in Appendix \ref{appendix: heartsteps trajectory model}.

\subsection{Synthetic Example}
\label{section: toy experiment}
The purpose of this experiment is to demonstrate that the trajectory model is able to correctly recover kernel compositions. We assume a toy environment where the goal is to perform 1-D regression on multiple users. There are $M_\mathrm{train} = 6$ total training users with $T_{m} = 6$  total time steps for all $m \in \mathbf{m}_\mathrm{train}$. The users are sampled from Gaussian Processes with one of two possible compositions: a $\mathrm{LIN} + \mathrm{PER}$ kernel or an $\mathrm{SE}$ kernel. To mimic the online setting, we reveal $[3, 4, 10, 20, 50, 100]$ new points at each time step. We use these inconsistently sized batches in order to emphasize the difference between the compositions found with small and large amounts of data. 
\begin{figure}[h]
    \centering
    \includegraphics[width = 0.8\linewidth]{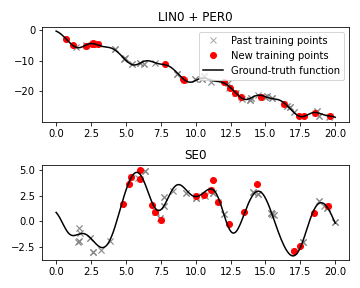}
    \caption{Example of a user drawn from a $\mathrm{LIN} + \mathrm{PER}$ GP (top) and a $\mathrm{SE}0$ GP (bottom) at $t = 5$. In an mHealth study, the y-axis (target) could represent the user's step count following an intervention. The x-axis could represent a $1$-dimensional state feature, such as temperature.}
    \label{fig:toy_experiment_example}
\end{figure}

\subsection{HeartSteps}
In this experiment, we evaluate the trajectory model on a real data set from an mHealth trial. HeartSteps V1 was a six week trial involving $37$ sedentary adults \cite{klasnja2019efficacy}. The goal of the HeartSteps app is to promote physical fitness (distal outcome) by maximizing the patient's step count (proximal outcome). The \emph{reward model} is a regression on the log-transformed $30$ minute step count following an intervention and is learned with the compositional GP. An intervention is a push notification, which also corresponds to the binary action space (to send or not to send). The raw state features involve $47$ covariates such as weather, temperature, the patient's location, time of day, and step count preceeding a decision point. A decision point is a fixed time throughout the day where the algorithm chooses to proceed with an intervention; it is a tuple $(\mathbf{s}, a, r)$ consisting of the state features, action, and outcome at a decision time, respectively. In Heartsteps there are $5$ decision points per day. Decision points form the data set $\mathcal{D}_{m, t} = \{(\mathbf{s}_n, a_n, r_n)\}_{n = 1}^{N_{m, t}}$ for a user $m$ at time $t$. To mimic the online setting, we split the data set for each user into chunks of $40$ points. Each user has a different number of total decision points, and as a result, will have a varying number of total time steps $T_m$. Our experiments use a subset of $21$ of the $37$ users that had a similar number of total (non-null) decision points with promising variation in step-count across decision points. A detailed data set summary is given in the appendix \ref{appendix: heartsteps data description}. 

\section{Results and Discussion}
\paragraph{The trajectory model adapts the kernel complexity to the amount of data.}
We expect the kernel for a user will grow in complexity (i.e. have more kernel components) as more data is observed. With the synthetic example in Section \ref{section: toy experiment}, it would be impossible to recover the periodic component of the $\mathrm{LIN} + \mathrm{PER}$ kernel for any user after recording only $3$ observations at $t=1$. Instead we expect a simpler kernel to emerge at $t = 1$, such as the $\mathrm{LIN}$ or $\mathrm{SE}$ kernel. At this point the simpler kernel is \emph{preferable} to prevent overfitting; this is reflected in the likelihood values in Appendix Table \ref{table: lml for data sizes}.

\begin{figure}[h]
    \centering
    \includegraphics[width = 0.5\linewidth]{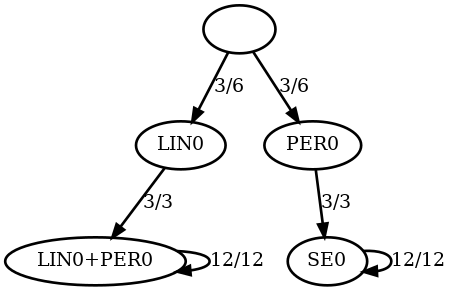}
    \caption{The trajectory model learned on training users. Edges represent a transition from $K_{t} \rightarrow K_{t + 1}$. Edges are labeled by the total number of times a kernel left a composition (denominator) to adopt the child node (numerator).}
    \label{fig:toy trajectory model}
\end{figure}

In Figure \ref{fig:toy trajectory model}, we visualize the kernel transitions which are learned on the $6$ training users from the toy experiment defined in Section \ref{section: toy experiment}. An analogous table of compositions found for the training users at each time step is shown in Appendix \ref{appendix: hs table compositions}. After the first time step, the users which have a ground truth $\mathrm{LIN} + \mathrm{PER}$ kernel vs. the $\mathrm{SE}$ kernel begin to diverge. However, the composition does not immediately become the ground-truth kernel, but rather, adopts a simpler intermediary when the data size is small. 


\paragraph{The trajectory model suggests appropriate kernel compositions on new data.}

The strength of the trajectory model is that it can perform kernel selection on previously unseen users without the need to retrain a selection algorithm.  
We report the result of predicting compositions with the Trajectory model from Fig. \ref{fig:toy trajectory model} on data from $M_\mathrm{test} = 100$ new users. The test users were created by sampling from a GP in the same manner as the training users, where $50$ users have a $\mathrm{LIN} + \mathrm{PER}$ kernel and the other $50$ a $\mathrm{SE}$ . More users transition from the simpler $\mathrm{LIN}$ and $\mathrm{PER}$ composition to the more complex  $\mathrm{LIN} + \mathrm{PER}$ and $\mathrm{SE}$ as the number of respective observations grows. Since our prediction procedure in Algorithm \ref{alg: trajectory prediction} allows for self-loops at every node, none of the users are required to move into more complex kernels but still do with the introduction of data. Finally, we report a confusion matrix of the predicted and true kernels at the last time step ($N_{m, 6} = 187$) in Appendix Figure \ref{fig:toy confusion matrix}.
\begin{figure}[t]
    \centering
    \includegraphics[width = 1\linewidth]{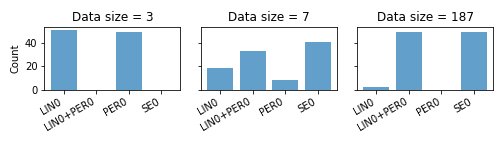}
    \caption{Count of compositions found across each of the $100$ test users for different data sizes. The number of users with complex compositions increases with the data size.}
    \label{fig:toy composition hist}
\end{figure}
\vspace{-0.5em}
\paragraph{ On Heartsteps data, the group level trajectory model has the same performance as the completely user level baselines -- with significant gains in prediction speed.}
In this experiment, we trained a Trajectory model on $M_{\mathrm{train}} = 15$ Heartsteps users to predict kernel compositions for $M_{\mathrm{test}} = 6$ new Heartsteps users. We compare our approach with Shrinkage \cite{tong2020shrinkage} and CKS \cite{duvenaud2013cks}, two completely user level baselines. 

\begin{figure}[t]
\centering
\includegraphics[width=1.\linewidth, height=170px]{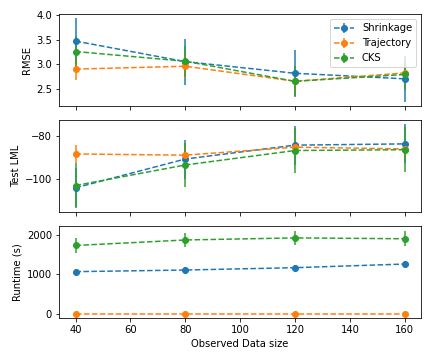}
\caption{Root Mean Square Error (top), test likelihood (middle), and runtime for prediction (bottom) of the kernel selection algorithms across Heartsteps users. Each metric represents the algorithm evaluated on test data at the next time step within a user. The error bars represent variability in the metrics across the users. The trajectory model has similar predictive performance and better run time compared to the baselines.}
\label{fig:hs comparisons}
\end{figure}

In Figure \ref{fig:hs comparisons} we compare the three selection algorithms with respect to RMSE, likelihood, and run time for predictions on the test users. All algorithms eventually select kernels with similar predictive performance. The trajectory model performs noticeably better at the first time step with around $40$ observed data points, potentially because it is able to leverage information from multiple users in order to predict the kernel for a test user with few observations. Since the trajectory model does not need to be updated, the predictive runtime after observing a new batch of data is constant.

\paragraph{The kernel transitions make sense for the mHealth application.}
\begin{figure*}[t]
    \centering
    \includegraphics[width = 0.7\linewidth]{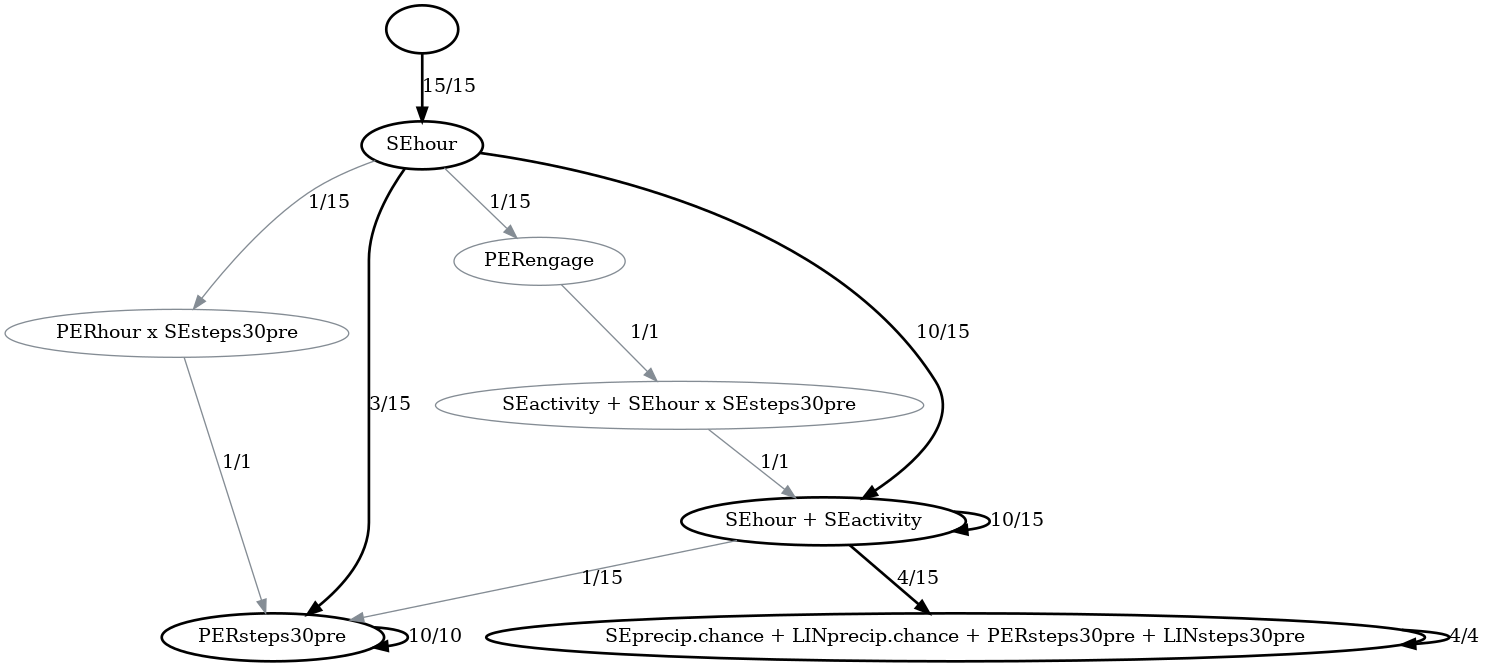}
    \caption{Structure of kernel trajectories found for the Heartsteps training users. Key compositions that were found for more than one user and/or time step are in bold.}
    \label{fig:heartsteps trajectories}
\end{figure*}
The structure of the Trajectory model itself is shown in Figure \ref{fig:heartsteps trajectories}. This model is \emph{transparent}; it is clear which kernel transitions were found by the model during training and which paths the model will take to select a kernel for new data. It is also \emph{interpretable} in that it is easy to identify how each kernel composition relates to the target. We identified the following key compositions: 
\begin{itemize}[leftmargin=*]
\itemsep 0em
    \item $\mathrm{SE}_\mathrm{hour}$: the user's step count will be similar to the step count we observed at a \emph{similar time}. The time is similar if it is within $3$ hours of the current time (lengthscale of $3.06$ hours). For example, the number of steps a user takes at $5$ pm will be similar to the number of steps they took in previous days around $2$ pm to $8$ pm. 
    \item $\mathrm{PER}_\mathrm{steps30pre}$: the user's step count $30$ minutes prior to the decision point is predictive of his future step count. The number of steps a user takes following $0$ preceding steps will be similar to the number of steps a user takes following $1465$ steps (period of $1465$ steps). This result is inconsistent with the results of prior analyses; 
    previously, we found that the more people walked prior to the decision point the more they walked after the decision point. This may be a sign of overfitting, discussed in Section \ref{section: discuss  overfitting}  
    \item $\mathrm{SE}_\mathrm{hour} + \mathrm{SE}_\mathrm{activity}$: the user's step count will be similar to the step count we observed at a \emph{similar time} (lengthscale $2.91$ hours). Independent of the time, it will also be similar to the step count we observed at a \emph{similar level of activity} (lengthscale $0.21$). The possible activities levels of activity are: still, active, and unknown. 
    \item $\mathrm{SE}_\mathrm{precip.chance} + \mathrm{LIN}_\mathrm{precip.chance} + \mathrm{PER}_\mathrm{steps30pre} + \mathrm{LIN}_\mathrm{steps30pre}$: the user's step count will be similar to the step count we observed at a \emph{similar time} (lengthscale $3.08$ hours). It will also be similar to the step count we observed at a \emph{similar chance of rain} (lengthscale $16\%$ chance of rain). The duplicitous kernels on the ``steps30pre" and ``precip.chance" covariates may indicate that the algorithm included the same covariate twice instead of increasing the amplitude hyperparameter for a single kernel. 
    \end{itemize}
    
These results align with our intuition that covariates such as activity level and preceding step count are important predictors for future step count and is consistent with the compositions found by other kernel selection methods. An example of this is shown in Appendix \ref{appendix: hs baseline compositions}, where all three selection methods included a periodic kernel on the ``steps30pre" covariate. Given the difficult nature of predicting step count from a user's context (imagine a user who walks sporadically throughout the day) more data may be necessary in order to isolate more interesting kernels from the high level of noise. The transitions within the model also make sense. For example, the $\mathrm{SE}_\mathrm{hour}$ composition appears at the top of the structure, at the earliest time step. A kernel on the ``hour" reappears further down the structure (at later time steps), in the  $\mathrm{SE}_\mathrm{hour} + \mathrm{SE}_\mathrm{activity}$ composition. 



\section{Discussion and Future Work} 
\paragraph{We may need to regularize the compositions.}
\label{section: discuss overfitting}
In some iterations of the stochastic optimization for our synthetic example (alternative iterations shown in \ref{appendix: toy model alternative}), the trajectory model resulted in a $\mathrm{LIN} + (\mathrm{PER} \times \mathrm{PER})$ or a $\mathrm{SE} \times \mathrm{PER}$ kernel as the final composition for the users. These compositions were multiplied by an extra $\mathrm{PER}$ kernel compared to the ground truth. Oftentimes, the trajectory models that contained these complicated kernels had a slightly higher likelihood than the structure in Fig. \ref{fig:toy trajectory model}, even though Fig. \ref{fig:toy trajectory model} contains the ground truth composition and may be preferable from an interpretability standpoint. 

In theory, the determinant term in the log marginal likelihood (Equation \ref{eq: gp log marginal likelihood}) acts as a complexity penalty for the model. We may need to consider a strategy to further regularize the composition and prevent overfitting. Previous approaches have used the Bayesian Information Criteria for model comparison \cite{duvenaud2013cks, kim2018scaling}, which includes a penalty to the composition based on the number of hyperparameters. 

\vspace{-0.5em}
\paragraph{Cost of training.}
The cost of training is the primary setback of the trajectory model. Since the goal of training is to associate \emph{cumulative} data sets to kernel compositions, 
optimization slows dramatically as the number of users and times steps increase. Future work will focus on decreasing the cost of training. This may be accomplished by alleviating the cost of GP matrix inversions and approximate inference for the Dirichlet Processes. 

\section{Conclusion}
In this paper we proposed a novel model for kernel selection for models in mobile health, which are rooted in the online, multi-user setting. Critically, our model avoids re-performing kernel selection by learning how kernels evolve under incremental data. We demonstrated that the trajectory model is able to recover compositions in a synthetic example and suggest useful compositions on new data.

Our paper explored the use of a compositional GP for learning a reward model in an mHealth app. To this end, we demonstrated that the Trajectory model selects kernels in a way that is transparent, decomposable, and makes sense with respect to the end application of predicting step counts. In future work we hope to apply the Trajectory model to the Heartsteps V2 data, which was collected from a longer study with more users. 

\paragraph{Acknowledgements}
Research reported in this work was supported by the National Institute Of Biomedical Imaging And Bioengineering and the Office of the Director of the National Institutes of Health under award number P41EB028242, the National Heart, Lung, and Blood Institute of the National Institutes of Health Under award number R01HL125440. The content is solely the responsibility of the authors and does not necessarily represent the official views of the National Institutes of Health. This material is also based upon work supported by the National Science Foundation Graduate Research Fellowship under Grant No. DGE1144152 and NSF CAREER 1750358. Any opinion, findings, and conclusions or recommendations expressed in this material are those of the authors and do not necessarily reflect the views of the National Science Foundation.
\bibliographystyle{icml2021}
\bibliography{main}

\onecolumn
\appendix
\section*{APPENDIX}
\section{Inference}
\label{appendix: mh proposer}
\subsection{Metropolis Hastings Proposal Distribution}
Let $K$ be the current kernel composition, which is a sum of $N_K$ kernels. Let $N$ be the total number of candidate kernels in the kernel pool, $\mathcal{K}$. A new composition $K'$ is proposed by randomly adding or removing a kernel from $K$.

\begin{align}
    \label{eq: mh probabilities}
    p(\text{action} = \text{add} | K) = 
    \begin{cases}
    0.3,  & \text{if } N_K < N \text{ and } N_K > 1\\
    1,              & \text{if } N_K \le 1\\
    0,  & \text{otherwise}
    \end{cases}\\
    p(\text{action} = \text{remove} | K) = 
    \begin{cases}
    0.7,  & \text{if } N_K < N \text{ and } N_K > 1\\
    1,              & \text{if } N_K = N\\
    0,  & \text{otherwise}
    \end{cases}
\end{align}

Once the action to add or remove a kernel is determine, we need to choose \emph{which} kernel to add or remove to the composition. If the action is to add, there is a $\frac{1}{N - N_K}$ chance of $K'$ because there are $N$ kernels you can add (from the kernel pool) that aren't already in the composition. If the action is to remove, there is a $\frac{1}{N_K}$ chance of $K'$ because you are removing one of the current kernels in the composition. 

\begin{align}
    \label{eq: mh probabilities}
    p(\tilde{K} | \text{action}, K) = 
    \begin{cases}
    \frac{1}{N - N_K},  & \text{if action } = \text{add} \\
    \frac{1}{N_K},              & \text{otherwise}
    \end{cases}
\end{align}

\subsection{Hyperparameter Distribution Updates}
\label{appendix: hyperparameter updates}

The distribution over hyperparameters is empirically updated based on the most likely hyperparameters to exist among the model's current compositions. For a given kernel composition $K$ and prior over hyperparameters $p_0(\boldsymbol \theta)$, the distribution over hyperparameters $p(\boldsymbol \theta | K)$ is updated as follows:

\begin{enumerate}
    \item Find the set of $\{\boldsymbol \theta^*\}$ that maximize $ \max \limits_{\boldsymbol \theta_{m, t}} p(y_{m, t} | X_{m, t}, K_{m, t} = K, \boldsymbol \theta_{m, t}) p_0(\boldsymbol \theta_{m, t})$. 
    \item $\hat p(\theta | K) = \frac{1}{N_K} \sum \delta_{\boldsymbol \theta^*_{m, t}}$ where $N_K$ is the number of assignments to $K$.
\end{enumerate}

\subsection{Prediction}
\label{appendix: hs prediction}
\begin{algorithm}[H]
\caption{Kernel prediction for data set $\mathcal{D}_{m, t+1}$}
\label{alg: trajectory prediction}
\SetAlgoLined
  \SetKwInOut{Input}{input}
  \SetKwInOut{Output}{output}
  \Input{Updated batch data $\mathcal{D}_{m, t + 1}$}
  \Output{Selection of $K_{m, t+1}$}
  $\mathbf{X}, \mathbf{y} = \mathcal{D}_{m, t + 1}$\;\\
  \eIf{t + 1 = 0} { //\text{New user}\\
    $K_\mathrm{prev} = ""$\; 
  }
  {
    $K_\mathrm{prev} = K_{m, t}$
  }
  
  $\mathbf{K} = \mathbf{K}^{(K_\mathrm{prev})} \cup \{K_{t - 1}\} $\;\\
  $K_{t + 1} = \max\limits_{K \in \mathbf{K}}  p(\mathbf{y} | \mathbf{X}, K)$\; 
 
\end{algorithm}

\section{Training Procedure for Kernel Selection Methods}
\begin{table}[H]
\label{table:train test}
    \centering
    \caption{Training procedure for each method. We used the optimization strategy which was most conducive to batch updates for the baselines. The procedure considers a mutually exclusive set of training ($\mathbf{m}_\mathrm{train}$) and testing ($\mathbf{m}_\mathrm{test}$) users.}
    \begin{tabular}{|m{0.3\linewidth}|m{0.3\linewidth}|m{0.3\linewidth}|}
        \hline
         CKS &  Shrinkage & Trajectory\\
         \hline
         Optimize model by restarting CKS for every $\mathcal{D}_{m, t}: m \in \mathbf{m}_\mathrm{test}$ and $t = 1, \ldots, T_m$
         & Optimize model by warm starting at the optima from previous time step for every $\mathcal{D}_{m, t}: m \in \mathbf{m}_\mathrm{test}$ and $t = 1, \ldots, T_m$
         & Optimize model over $\mathcal{D}_{m, t}: m \in \mathbf{m}_\mathrm{train}$ and $t = 1, \ldots, T_m$\\
         \hline
    \end{tabular}
    \label{table: train baselines}
\end{table}

\section{Synthetic Experiment}
\subsection{Candidate Kernel Pool}
\label{appendix: toy pool}
Since this experiment is $1$ dimensional, the candidate kernel pool for the shrinkage and trajectory models is the polynomial expansion of all kernel functions up to degree $2$. This includes the following $8$ models: 
\begin{enumerate}
\itemsep 0em
    \item $\mathrm{LIN}$
    \item $\mathrm{PER}$
    \item $\mathrm{SE}$
    \item $\mathrm{LIN} \times \mathrm{LIN}$
    \item $\mathrm{LIN} \times \mathrm{PER}$
    \item $\mathrm{LIN} \times \mathrm{SE}$
    \item $\mathrm{PER} \times \mathrm{SE}$
    \item $\mathrm{PER} \times \mathrm{PER}$
\end{enumerate}

Note that the $\mathrm{SE} \times \mathrm{SE}$ kernel is not included because this is simply the $\mathrm{SE}$ kernel. 

\subsection{Priors}
\label{appendix: toy priors}
The following weak priors are placed on the hyperparameters of each kernel in the synthetic experiment, by type: 
\begin{itemize}
\itemsep 0em
    \item Lengthscale $p(\ell | K) = \text{LogNormal}(\mu = 0, \sigma^2 = 0.5)$
    \item Amplitude $p(\sigma^2_f | K) = \text{LogNormal}(\mu = 1, \sigma^2 = 0.5)$
    \item Period $p(p | K) = \text{LogNormal}(\mu = 1, \sigma^2 = 0.5)$
    \item Location (for linear kernel function) $p(c | K) = \text{Normal}(\mu = 0, \sigma^2 = 0.1)$
    \item Noise variance $p(\sigma^2 | K) = \text{LogNormal}(\mu = 0, \sigma^2 = 0.5)$
\end{itemize}

Each Dirichlet Process must also include a base distribution $H_0$ over possible kernel compositions. In the toy experiment, the base distribution places an independent Bernoulli prior on each kernel in the candidate kernel pool from Appendix \ref{appendix: toy pool}. Let $N_K$ denote the number of kernels that are being multiplied in $K$. For example, $N_{\mathrm{LIN} \times \mathrm{LIN}} = 2$ and $N_{\mathrm{LIN}} = 1$. The probability that each kernel $K$ from the candidate kernel pool is included in the composition is as follows: 
\begin{equation}
    p(K) = 
    \begin{cases}
    0.1,  & \text{if } N_K  = 1\\
    0.25,  & \text{otherwise}
    \end{cases}\\
\end{equation}
we place greater probability for simpler (1 term) kernels in order to prioritize the compositions to include simpler kernels. Finally we set $\alpha = 1$. 

\subsection{Example of potential for overfitting}
\begin{table}[H]
    \centering
    \caption{Log likelihood for ground-truth kernel ($\mathrm{LIN} + \mathrm{PER}$) vs. simpler kernel $\mathrm{LIN}$ for small ($3$) and large $187$ number of data points. Notice that the ground-truth composition has lower likelihood when the data set is small due to overfitting.}
    \begin{tabular}{c|c|c|}
         & Data size $3$& Data size $187$\\ \hline
         $\mathrm{LIN}$ & $-5.86$ & $-114.67$\\
         $\mathrm{LIN} + \mathrm{PER}$ &$-6.22$  & $-84.44$\\ 
    \end{tabular}
    \label{table: lml for data sizes}
\end{table}

\subsection{Synthetic Experiment Models}
\label{appendix: toy model alternative}
\begin{figure}[h]
\centering
\includegraphics[width=0.4\linewidth]{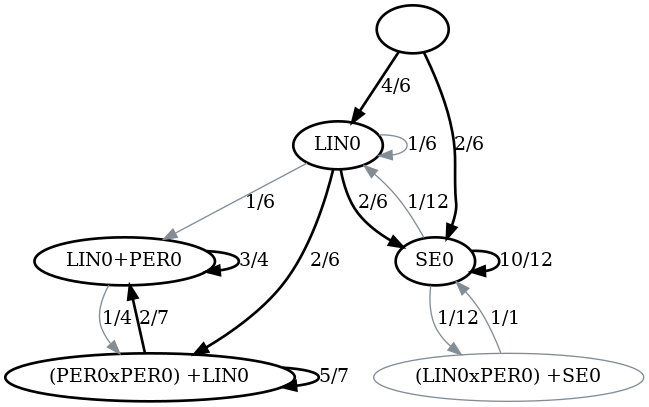}
\includegraphics[width=0.4\linewidth]{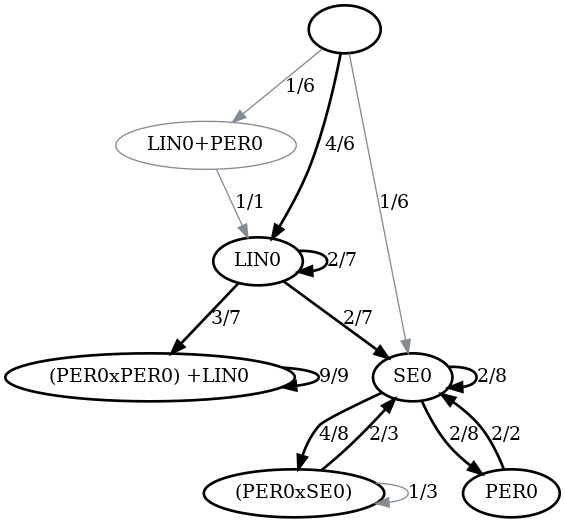}
\includegraphics[width=0.7\linewidth]{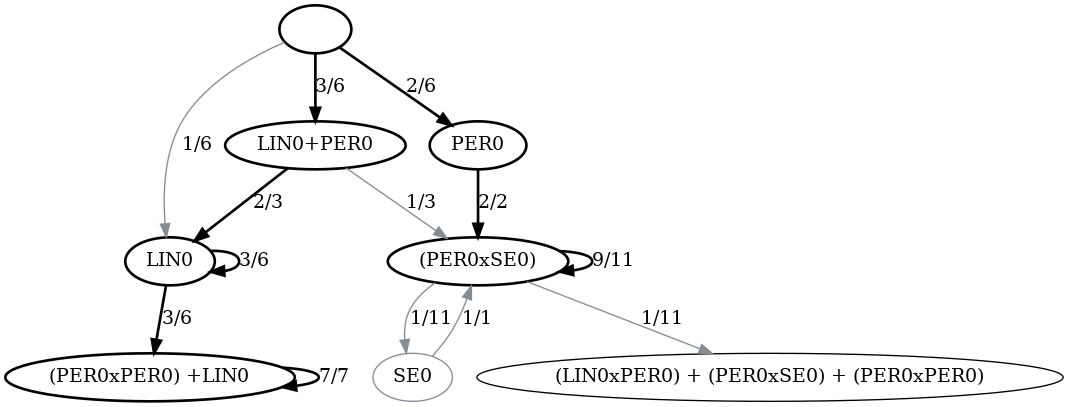}
\caption{Alternate trajectory model structures resulting from different iterations of the optimization process. All structures are within $55$ log likelihood values of one another, and the model in Figure \ref{fig:heartsteps trajectories}}
\end{figure}

\label{appendix: hs table compositions}
\begin{table}[h]
\caption{The kernel compositions learned for each training user and time step in the synthetic experiment, corresponding to the structure shown in Figure \ref{fig:toy trajectory model}.}
    \centering
    \begin{tabular}{c|c | c | c | c | c | c}
         $m$ & $t = 1$ & $t = 2$ & $t = 3$ & $t = 4$ & $t = 5$  & \textbf{Ground Truth}\\ \hline
         $1$ & $\mathrm{LIN}$ & $\mathrm{LIN} + \mathrm{PER}$& $\mathrm{LIN} + \mathrm{PER}$ & $\mathrm{LIN} + \mathrm{PER}$ & $\mathrm{LIN} + \mathrm{PER}$ & $\mathrm{LIN} + \mathrm{PER}$\\
         $2$ & $\mathrm{LIN}$ & $\mathrm{LIN} + \mathrm{PER}$& $\mathrm{LIN} + \mathrm{PER}$ & $\mathrm{LIN} + \mathrm{PER}$ & $\mathrm{LIN} + \mathrm{PER}$ & $\mathrm{LIN} + \mathrm{PER}$\\
         $3$ & $\mathrm{LIN}$ & $\mathrm{LIN} + \mathrm{PER}$& $\mathrm{LIN} + \mathrm{PER}$ & $\mathrm{LIN} + \mathrm{PER}$ & $\mathrm{LIN} + \mathrm{PER}$ & $\mathrm{LIN} + \mathrm{PER}$\\
         $4$ & $\mathrm{PER}$ & $\mathrm{SE}$& $\mathrm{SE}$ & $\mathrm{SE}$ & $\mathrm{SE}$ & $\mathrm{SE}   $\\
         $5$ & $\mathrm{PER}$ & $\mathrm{SE}$& $\mathrm{SE}$ & $\mathrm{SE}$ & $\mathrm{SE}$ & $\mathrm{SE}$\\
         $6$ & $\mathrm{PER}$ & $\mathrm{SE}$& $\mathrm{SE}$ & $\mathrm{SE}$ & $\mathrm{SE}$ & $\mathrm{SE}$\\
    \end{tabular}
    \label{table: toy compositions}
\end{table}

\subsection{Confusion Matrix}
\begin{figure}[H]
    \centering
    \includegraphics[width = 0.25\linewidth]{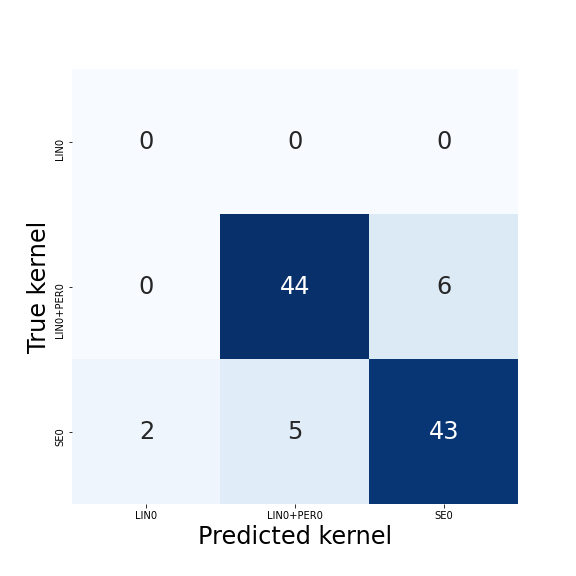}
    \caption{Confusion matrix of true vs. predicted composition for all $100$ test users at final time step ($N_{m, 6} = 187$). A majority of users are correctly binned into the ground-truth composition by the last time step. 
    }
    \label{fig:toy confusion matrix}
\end{figure}

\section{Heartsteps Experiment}
\subsection{Dataset Description}
\label{appendix: heartsteps data description}
The prediction target is the log transformed $30$ minute step count following a decision point. We remove all decision points for which the target is missing or unrecorded (due to connection errors, hardware errors). 

The covariates for all data sets are min-max scaled to the range of $0$ to $1$. To be consistent with the training data available to each baseline, the data is scaled according to the \emph{user level} data for the shrinkage and CKS models. The data is scaled according to the \emph{group level} training data for the trajectory model.  

{\renewcommand{\arraystretch}{1.3} 
\begin{table}[h]
\begin{center}
\caption{Overview of Heartsteps V1 data} 
\begin{tabular}{| l  l |  }
\hline
\textbf{Num. features} & $47$\\
\textbf{Num. expert mapped features} & $7$\\
\textbf{Num. candidate kernels} & $68$\\
\textbf{Num. train users} & $M_\mathrm{train} = 15$\\
\textbf{Num. test users} & $M_{\mathrm{test}} = 6$\\
\textbf{Num. time steps per train user} & $T_{m_\mathrm{train}} = [3, 5, 4, 4, 5, 4, 2, 4, 5, 4, 4, 4, 5, 5, 4]$\\
\textbf{Num. points per train user} & $N_{m_\mathrm{train}} = [110, 178, 147, 142, 167, 133, 68, 147, 175, 150, 152, 156, 173, 190, 150]$\\
\textbf{Num. time steps per test user} & $N_{m_\mathrm{test}} = [5, 5, 5, 5, 5, 4]$\\
\textbf{Num. points steps per test user} & $T_{m_\mathrm{test}} = [169, 171, 182, 179, 164, 137]$\\

\hline
\end{tabular}
\end{center}
\end{table}
}

\subsection{Reward Model}
\label{appendix: hs reward model}
In Heartsteps the reward model represents a user specific mapping from states and actions to step counts. The RL bandit's actions are determined by the predicted impact of the intervention on the user. The reward model for user $m$ can be broken down into a separate baseline and advantage function. The advantage function represents the effect of taking an action. To see this, note that $r_m(\mathbf{s}, 1)-r_m(\mathbf{s}, 0)=f_m^\mathrm{advantage}$ in \ref{eq: reward model template}. Conversely, the baseline function captures the baseline effect of the state on the step count, when no notification is sent. 

\begin{align}
\label{eq: reward model template}
  r_m(\mathbf{s}, a) = f^\mathrm{baseline}_m(\mathbf{s}) + a * f^\mathrm{advantage}_m(\mathbf{s}) + \epsilon  
\end{align}

In this experiment, we assume that $f_\mathrm{advantage}$ is a linear function and $f_\mathrm{baseline}$ is a compositional GP whose kernel is determined by the kernel selection algorithm. The resulting model is a Gaussian Process: 
\begin{align}
\label{eq: reward model GP}
r_m(\mathbf{s}, a) \sim GP(c_m, k_m^\mathrm{baseline} + k_m^\mathrm{advantage}+ \sigma_m^2 I)
\end{align}
where $c_m$ is a constant mean function and $k_m^\mathrm{advantage}(\mathbf{x}, \mathbf{x}') = \mathbf{x}^\top \Sigma\mathbf{x}'$. The vector $\mathbf{x}$ is a concatenation of expert recommend state and action. That is, $\mathbf{x} = [1, a * \phi(\mathbf{s})]$ where $\phi$ is the expert mapping of the raw state features (usually a summary or a subset of features) and $1$ is a bias term.

\subsection{Candidate Kernel Pool}
\label{appendix: hs pool}
The candidate kernel pool for both the shrinkage and trajectory models consists of simple, lower-order base kernels. It includes each kernel function applied to each covariate in the data. The noise level is high in digital health and thus the local structure that would be detected by products of kernels is harder to detect. For this reason, we only include a small selection of multiplicative kernels based on our preliminary exploration of the data. 

\subsection{Priors}
\label{appendix: hs priors}
The following priors are placed on the hyperparameters of the kernel compositions. These priors are consistent for all three kernel selection methods. They are different from the priors placed on the synthetic experiment from \ref{appendix: toy priors} because they are placed on the scaled Heartsteps data. 

\begin{itemize}
\itemsep 0em
    \item Lengthscale $p(\ell | K) = \text{LogNormal}(\mu = -1, \sigma^2 = 0.75)$
    \item Amplitude $p(\sigma^2_f | K) = \text{LogNormal}(\mu = 0.5, \sigma^2 = 0.75)$
    \item Period $p(p | K) = \text{LogNormal}(\mu = -1, \sigma^2 = 0.75)$
    \item Location (for linear kernel function) $p(c | K) = \text{Normal}(\mu = 0, \sigma^2 = 0.1)$
    \item Noise variance $p(\sigma^2 | K) = \text{LogNormal}(\mu = 1, \sigma^2 = 0.75)$
\end{itemize}

Like in the synthetic case, the base distribution $H_0$ places a bernoulli distribution over each allowed kernel and higher probability is assigned to simpler (non-multiplicative) kernels. We also placed higher likelihood on certain kernels, based on preliminary analysis (of validation data exclusive from the training users). This set of kernels is $\mathbf{K}_\mathrm{preliminary}= \{\mathrm{LIN}_\mathrm{min}, \mathrm{SE}_\mathrm{steps30pre}, \mathrm{PER}_\mathrm{hour}, \mathrm{SE}_\mathrm{hour}, \mathrm{SE}_\mathrm{hour}, \mathrm{SE}_\mathrm{temperature}, \mathrm{PER}_\mathrm{temperature}, \mathrm{SE}_\mathrm{weather}, \mathrm{PER}_\mathrm{yday}, \mathrm{SE}_\mathrm{yday}, \mathrm{SE}_\mathrm{hour} \times \mathrm{SE}_\mathrm{weather}, \mathrm{SE}_\mathrm{hour} \times \mathrm{SE}_\mathrm{temperature}\}$. The probability of each kernel component is then: 
\begin{equation}
    p(K) = 
    \begin{cases}
    0.1,  & \text{if } K = \mathrm{PER}_\mathrm{steps30pre}\\
    0.05 & \text{if } K \in  \mathbf{K}_\mathrm{preliminary}\\
    0.03 & \text{if } N_K  = 1\\
    0.01 & \text{otherwise}\\
    \end{cases}\\
\end{equation}
Finally we set $\alpha = 1$. 

\subsection{Kernel Compositions found for Baselines}
\label{appendix: hs baseline compositions}
The following is a comparison of the kernel compositions found at each time step for a training user by all three selection algorithms. 
\begin{table}[H]
\centering
\caption{Comparison of kernels selected by each method for a Heartsteps training user.}
\begin{tabular}{|c|c|c|c|}
    \hline
    $t$ & Trajectory (ours) & Shrinkage \cite{tong2020shrinkage} & CKS \cite{duvenaud2013cks} \\
    \hline
    1 
    & $\mathrm{PER}_\mathrm{hour}$ 
    & $\mathrm{None}$ 
    & $\mathrm{SE}_\mathrm{steps30pre}$\\
    
    2 
    & $\mathrm{PER}_\mathrm{steps30pre}$
    & $\mathrm{PER}_\mathrm{steps30pre}$ 
    & $\mathrm{SE}_\mathrm{steps30pre} \times \mathrm{LIN}_\mathrm{day\ of\ month}$\\
    
    3 
    & $\mathrm{PER}_\mathrm{steps30pre}$
    & $\mathrm{PER}_\mathrm{steps30pre}$
    & $\mathrm{SE}_{steps30pre}$\\
    
    4
    & $\mathrm{PER}_\mathrm{steps30pre}$ 
    & $\mathrm{PER}_\mathrm{steps30pre}$
    & $\mathrm{SE}_{steps30pre}$\\
    
    5 
    & $\mathrm{PER}_\mathrm{steps30pre}$
    & $\mathrm{PER}_\mathrm{steps30pre}$
    & $\mathrm{PER}_\mathrm{steps30pre}$\\
    \hline
\end{tabular}
\label{table: hs baseline compositions}
\end{table}

\subsection{Heartsteps Trajectory Models}
\label{appendix: heartsteps trajectory model}
\begin{figure}[H]
\centering
\includegraphics[width=0.6\linewidth]{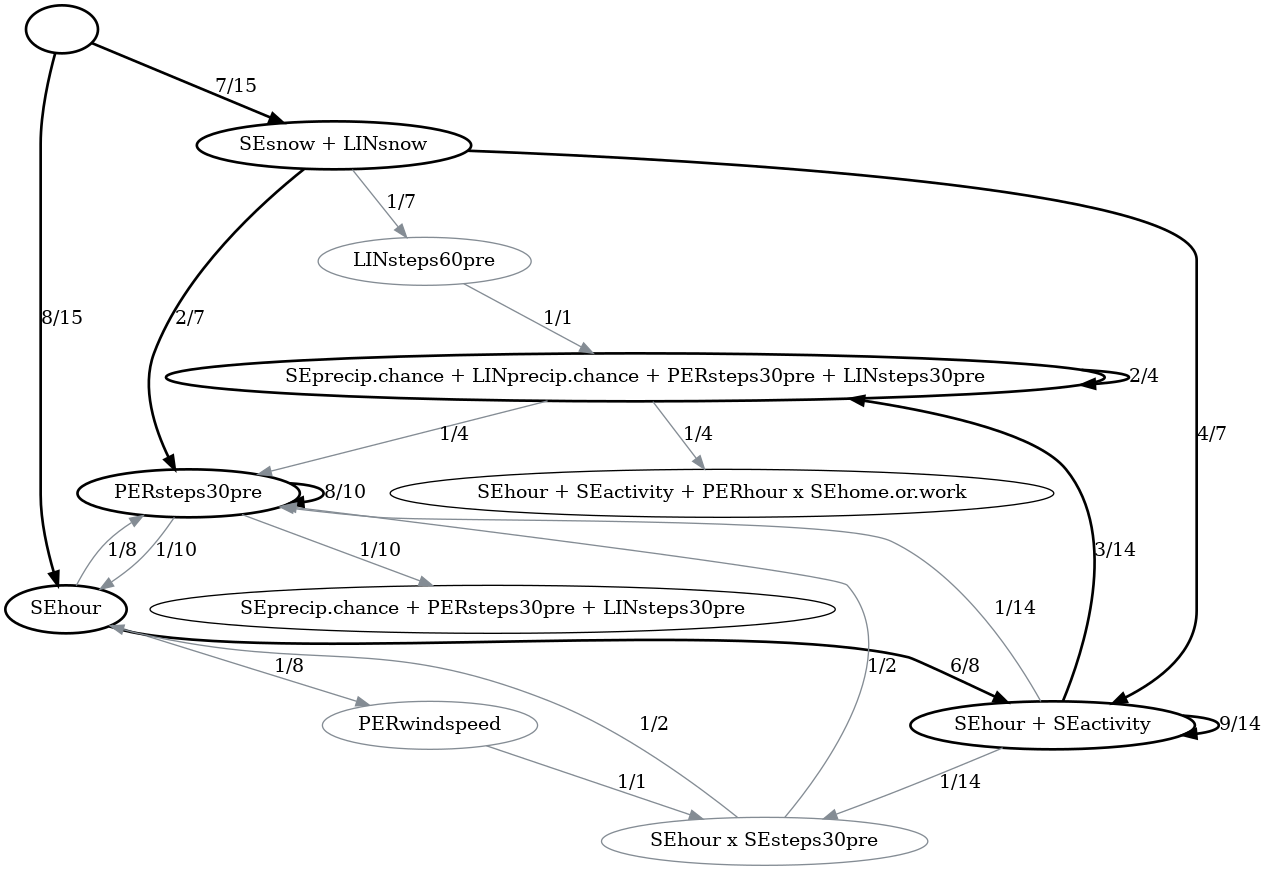}
\includegraphics[width=0.8\linewidth]{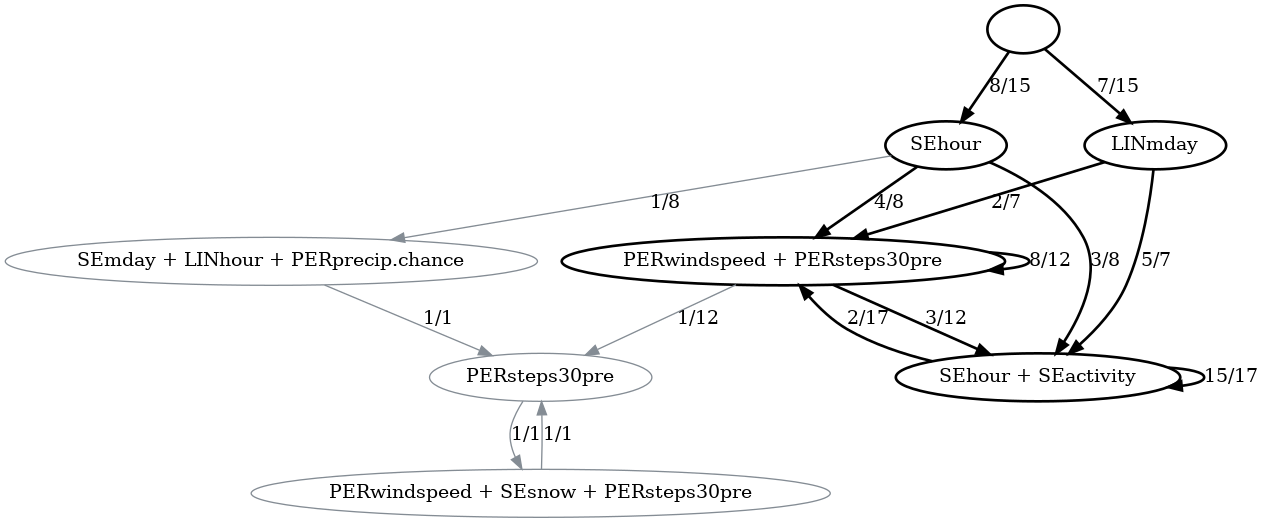}
\includegraphics[width=0.8\linewidth]{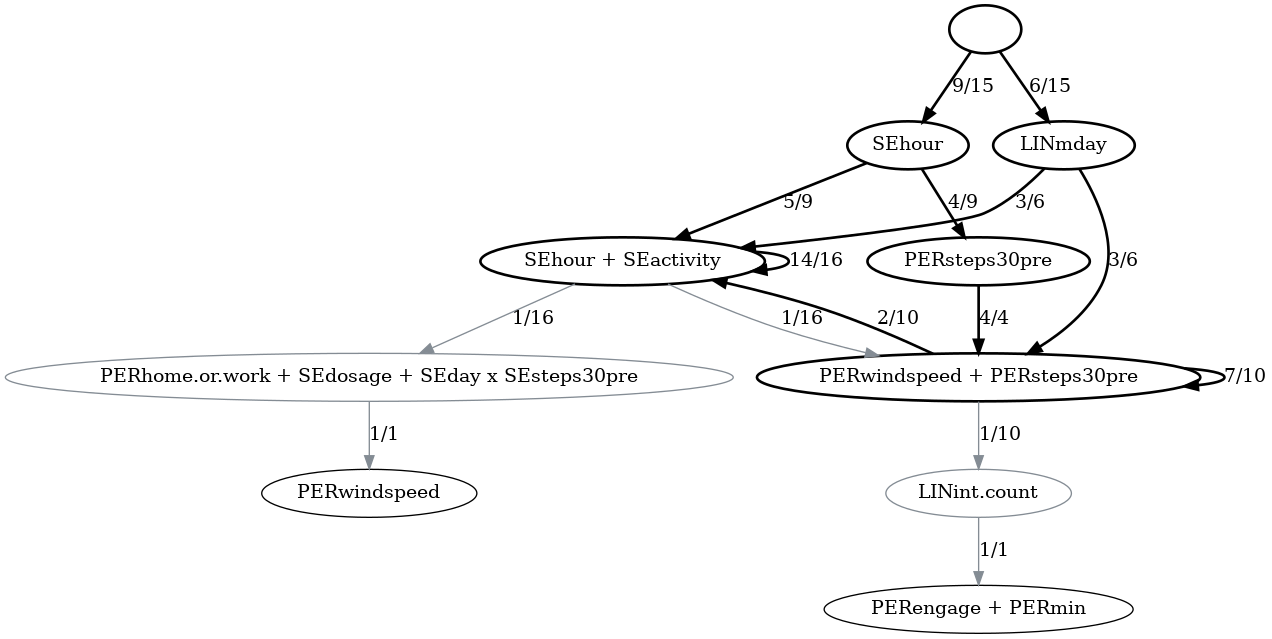}
\caption{Alternate trajectory model structures resulting from different iterations of the optimization process for Heartsteps experiment. All structures are within $24$ log likelihood values of one another, and the model in Figure \ref{fig:heartsteps trajectories}.}
\end{figure}

\end{document}